\documentclass[letterpaper, 10 pt, journal, twoside]{IEEEtran}

\usepackage{amsmath}
\usepackage{amssymb}
\usepackage{bm}
\usepackage{graphicx}
\usepackage{subcaption}
\usepackage[skip=0pt,labelfont=bf]{caption}
\usepackage{wrapfig}
\usepackage{adjustbox}
\usepackage{pgfplots}
\usetikzlibrary{patterns}
\usepackage{color,soul}
\usepackage{amssymb}
\usepackage{url}
\usepackage{breqn}
\usepackage[top=57pt, left=48pt, right=48pt, bottom=43pt]{geometry}
\usepackage{cite} 
\usepackage{multirow}
\usepackage{fixme}
\fxsetup{status=draft, theme=color}
\definecolor{fxtarget}{rgb}{0.8000,0.0000,0.0000}
\definecolor{fxnote}{rgb}{0.0000,0.0000,0.8000}

\definecolor{orange}{RGB}{241,163,64}
\definecolor{white}{RGB}{247,247,247}
\definecolor{purple}{RGB}{153,142,195}

\begin{document}
	
\title{\LARGE \bf
Multi-Fingered Grasp Planning via Inference in Deep Neural Networks}





\author{Qingkai Lu$^{1}$, Mark Van der Merwe$^{1}$, Balakumar Sundaralingam$^{1}$ and Tucker Hermans$^{1}$%
	\thanks{Q.~Lu and B.~Sundaralingam were supported in part by NSF Award 1846341.}
	\thanks{$^{1}$Qingkai Lu, Mark Van der Merwe, Balakumar Sundaralingam and Tucker Hermans are with the School of Computing and the Robotics Center, University of Utah, Salt Lake City, UT 84112, USA.
		{\tt\footnotesize qklu@cs.utah.edu; mark.vandermerwe@utah.edu; bala@cs.utah.edu; thermans@cs.utah.edu}}%
}

\maketitle

\begin{abstract}
 We propose a novel approach to multi-fingered grasp planning leveraging learned deep neural network models.
 We train a voxel-based 3D convolutional neural network to predict grasp success probability as a function of both visual information of an object and grasp configuration. We can then formulate grasp planning as inferring the grasp configuration which maximizes the probability of grasp success. In addition, we learn a prior over grasp configurations as a mixture density network conditioned on our voxel-based object representation.

 We show that this object conditional prior improves grasp inference when used with the learned grasp success prediction network when compared to a learned, object-agnostic prior, or an uninformed uniform prior. Our work is the first to directly plan high quality multi-fingered grasps in configuration space using a deep neural network without the need of an external planner. We validate our inference method performing multi-finger grasping on a physical robot. Our experimental results show that our planning method outperforms existing grasp planning methods for neural networks.
\end{abstract}




\section{Introduction and Motivation}
\label{sec:intro}
Learning-based approaches to grasping~\cite{saxena2006learning,Saxena-aaai2008,lenz2015deep,pinto2016supersizing,Kopicki2016, mousavian20196, wu2019pixel, liu2019generating} have become a popular alternative to geometric~\cite{sahbani2012overview, bohg2014data, ciocarlie2007dexterous, dragiev2011gaussian} and model-based planning~\cite{Grupen1991,Murray1994} over the past decade. In particular grasp learning has shown to generalize well to previously unseen objects where only partial-view visual information is available. Moreover by having a robot attempt and validate its own grasps in a self-supervised manner~\cite{pinto2016supersizing} we remove the need for humans to guess at what grasps would be successful~\cite{saxena2006learning} or for performing complex mappings from human grasps to robot grasps.

More recently, researchers have looked to capitalize on the success of deep neural networks to improve grasp learning. Broadly speaking deep neural network methods for grasp learning can be split into two approaches: predicting grasp success for the object visual information represented by an image patch or a point cloud associated with a gripper configuration~\cite{lenz2015deep, gualtieri2016high,pinto2016supersizing,levine2016learning,mahler2017dex,johns2016deep,varley2015generating,mousavian20196} and directly predicting a grasp configuration from the object visual information represented by an image or image patch using regression~\cite{redmon2015real, kumra2016robotic, veres2017modeling, liu2019generating}.
While these deep learning approaches have shown impressive performance for parallel jaw grippers (e.g.~\cite{pinto2016supersizing}) relatively little work has focused on the more difficult problem of multi-fingered grasping~\cite{varley2015generating,veres2017modeling,kappler2015leveraging,liu2019generating,wu2019pixel}. We believe two primary difficulties restrict the use of deep learning for multi-fingered grasping (1) the grasping input representation in neural networks and (2) the reliance on external planners for generating candidate grasps.

The first difficulty arises from (1) the great variability in object shapes a robot may encounter during deployment and (2) the high dimensionality of multi-fingered grasp configurations. As such, grasp learning requires representations that can efficiently, in terms of data and computation, encode both the geometry of the object necessary and the grasp configuration to predict grasp success. The second issue arises from the increase in search complexity for planning multi-fingered grasps compared to grasping with parallel jaw grippers. As such we desire an efficient inference procedure faster than typical model-based planners, which does not suffer from the limiting assumptions and bias present in such human-designed planners.


In order to combat these two problems, we propose an alternative approach to grasp planning with deep neural networks, where we directly use the learned network for planning. In our work, we train a network to predict grasp success, given an object and grasp configuration representation. However, unlike currently employed sampling methods, we perform a continuous optimization over the grasp configuration, guiding the updates with the network.


In addition to giving the neural network the object visual information represented by an RGB-D image or a voxel-grid as input, we also provide the grasp configuration parameters in the form of finger preshape joint angles and palm pose. However, we formulate the learning and inference problem in a general form that would allow other grasp representations to easily be used instead (e.g. joint angles of in-contact fingers, desired finger contact locations on the object surface, etc.).

Once trained, given the object visual representation, we perform inference over the grasp configuration parameters in order to maximize the probability of grasp success learned by our convolutional neural network (CNN). We perform this probabilistic inference as a direct optimization over the grasp configuration, which leverages the efficient computation of gradients in neural networks, while ensuring joint angles remain within their limits.
Thus, our approach can quickly plan reliable multi-fingered grasps given an image of an object and an initial grasp configuration.

This articles makes the following contributions over our previous work~\cite{lu2017grasp}:
\begin{itemize}
\item We propose a new grasp model for grasp learning and inference, including a voxel-based 3D convolutional neural network (CNN) for grasp success probability prediction that encodes the object 3D geometry well.
\item We present a novel mixture-density network (MDN) to model a grasp configuration prior conditioned on the observed object, which removes the need for an external grasp planner to initialize our grasp optimization.
\item Our grasp planner generates both successful side and overhead grasps on the real robot, while our grasp planner in~\cite{lu2017grasp} only generates successful side grasps on the real-robot.
\item We train our grasp model using more than seven times the data of~\cite{lu2017grasp}.
\item We perform real-robot grasp experiments on more objects than~\cite{lu2017grasp}.
\item In total this creates a grasp planner that achieves a higher success rate than the our previous planner in~\cite{lu2017grasp}.
\end{itemize}

Our planner offers a number of benefits over previous deep-learning approaches to multi-fingered grasping. Kappler and colleagues~\cite{kappler2015leveraging} learn to predict if a given palm pose will be successful for multi-fingered grasps using a fixed preshape and perform planning by evaluating a number of sampled grasp poses. Varley et al.~\cite{varley2015generating} present a deep learning approach to effectively predict a grasp quality metric for multi-fingered grasps, but rely on an external grasp planner to provide candidate grasps.
In contrast, our method learns to predict grasp success as a function of both the palm location and preshape configuration and plans grasps directly using the learned network. Saxena et al.~\cite{Saxena-aaai2008} also perform grasp planning as inference using learned probabilistic models; however they use separate classifiers for both the image and range data, using hand selected models instead of a unified deep model. Zhou and Hauser~\cite{zhou6dof} concurrently propose a similar optimization-based grasp planning approach to ours using a similar CNN architecture. In contrast to our work, they do not interpret planning as probabilistic inference; they optimize only for hand pose, ignoring hand joint configurations; and they validate only in simulation.

Veres et al.~\cite{veres2017modeling} train a conditional variational auto-encoder (CVAE) deep network to predict the contact locations and normals for a multi-fingered grasp given an RGB-D image of an object. In order to perform grasping an external inverse kinematics solver must be used for the hand to try and reach the desired contact poses as best as possible. Liu et al.~\cite{liu2019generating} train a 3D voxel CNN to directly predict the multi-finger grasp configuration. Implicit in such regression methods as proposed in~\cite{veres2017modeling, liu2019generating} lies the assumption that there exists a unique best grasp for a given object view.
In contrast, our method can plan multiple successful grasps for a given object using different initial configurations with associated high confidence prior to execution. This offers the robot the option of selecting a grasp best suited for its current task. Additionally, we show that our classification-based network can effectively learn with a smaller dataset compared with a regression network, which can not leverage negative grasp examples.

We formulated multi-fingered grasp planning as probabilistic inference in a learned deep neural network without a prior over grasp configuration in our previous work~\cite{lu2017grasp}. Our multi-channel deep neural network in~\cite{lu2017grasp} took a grasp configuration and RGB-D image grasp patch as inputs and predicts as output the probability of grasp success. Our planning algorithm generally achieved higher grasp success rates compared with sampling-based and regression approaches currently used for grasping with neural networks.

We explored a probabilistic graphical model for grasp learning and planning over grasp type and grasp configuration for a given object in our previous work~\cite{lu2019grasp}. We used a data-driven Gaussian mixture model (GMM) prior independent of the object to constrain the inference to not stray into areas far from grasp configurations observed at training time, where we have little evidence to support grasp success predictions. The grasping experiment results in~\cite{lu2019grasp} demonstrated the benefit of a data-driven prior for grasp inference.
In this article, we propose an object conditional prior modeled as a mixture density network (MDN)~\cite{bishop1994mixture}. Our MDN prior models the grasp configuration distribution based on the geometry of the object of interest. Our real-robot experiments show grasp inference with the MDN object conditional prior outperforms grasp inference with the GMM object independent prior.
We trained a logistic regression classifier to predict the grasp success probability on a small data-set with $120$ grasps in~\cite{lu2019grasp}. In this article, we train a voxel-based 3D CNN to predict the grasp success probability on a larger data-set containing $10,811$ grasp attempts.


In the next section we provide a formal description of our grasp planning approach. We follow this in Section~\ref{sec:grasp_learning} with an overview of our approach to multi-fingered grasp learning and the novel voxel-based 3D CNN architectures for predicting grasp success. We then give a thorough account of our experiments and results in Section~\ref{sec:exp}. We conclude with a brief discussion in Section~\ref{sec:conclusions}.


\section{Grasp Planning as Probabilistic Inference}
\label{sec:grasp_inference}

Following~\cite{ciocarlie2007dexterous} we define the grasp planning problem as finding a grasp preshape configuration. 
In our case, the grasp configuration vector is composed of the palm pose in the object reference frame and the hand’s preshape joint angles that define the shape of the hand prior to closing the hand.
In order to make the grasp inference agnostic to object poses, we put the palm pose in the object reference frame for learning and inference.
After finding the grasp preshape configuration, the robot moves to this preshape and runs a controller to close the hand forming the grasp on the object. We explain the specific joints used for defining the preshape and how the grasp controller works for our experiments in Section~\ref{subsec:data_collection}.
We focus on scenarios where a single, isolated object of interest is present in the scene. Importantly, we assume no explicit knowledge of the object beyond a single camera sensor reading of it in its current pose.
The problem we address states, given such a grasp scenario, plan a grasp preshape configuration that allows the robot to successfully grasp and lift the object without dropping it.

Given the learned model parameters, \(\bm W\) and \(\bm\Phi\), along with the visual representation, \(z\), associated with an observed object of interest, our goal is to infer the grasp configuration parameters, \(\bm\theta\), that maximize the posterior probability of grasp success \(Y=1\). Here \(Y\) defines a random Boolean variable with 0 meaning failure and 1 meaning success.
We can thus formalize grasp planning as a \emph{maximum a posteriori} (MAP) inference problem:
\begin{flalign}
\underset{\boldsymbol{\theta}}{\text{argmin}}\hspace{15pt}&-\log p(\boldsymbol{\theta}| Y=1, z, \bm{W}, \bm\Phi) \\
\text{subject to}\hspace{8pt}&{\bm\theta}_{\texttt{min}} \preceq \bm\theta \preceq {\bm\theta}_{\texttt{max}}
\label{eq:inf_obj}
\end{flalign}
We constrain the grasp configuration parameters to obey the joint limits of the robot hand in Eq.~\ref{eq:inf_obj}.

We define the grasp success likelihood \(p(Y=1 | \boldsymbol{\theta}, z, \bm{W})\) to be a deep neural network. \(\bm{W}\) represents the neural network parameters. The deep neural network predicts the probability of grasp success, \(Y\), as a function of the visual representation of the object of interest, \(z\), and hand configuration, \(\theta\).\footnote{We previously represented the object of interest as an RGB-D image patch in~\cite{lu2017grasp}. In this article, we use a voxel-grid to represent the object of interest.} We describe the details of our neural network classifier in Section~\ref{subsec:voxel_classifier}.

We present three different ways to model the prior over the grasp configuration \(\theta\). In each case \(\bm\Phi\) represents the associated parameters of the prior distribution.
In the first approach we assume a uniform prior over valid grasp configurations, resulting in the grasp success posterior probability being proportional to the likelihood as shown in Eq.~\ref{eq:inf_constraint}:
\begin{equation}
  p(\boldsymbol{\theta} | Y=1, z, \bm{W}, \bm\Phi) \propto p(Y=1 | \boldsymbol{\theta}, z, \bm{W}) \label{eq:inf_constraint}
\end{equation}
This prior requires all grasp parameters to be bounded to prevent the inference straying way from the training evidence. This approach was used in our initial work~\cite{lu2017grasp}. It is trivial to bound the preshape joint angles using the robot hand joint limits. However,  it requires heuristic bounds to be manually designed for the hand palm pose in Cartesian space.

The second prior we examine defines a prior over grasp configurations to encode preferred grasp configurations independent of the observed object, which gives the following posterior:
\begin{equation}
  p(\boldsymbol{\theta} | Y=1, z, \bm{W}, \bm\Phi) \propto p(Y=1 | \boldsymbol{\theta}, z, \bm{W}) p(\boldsymbol{\theta} | \bm\Phi) \label{eq:inf_GMM}
\end{equation}
In practice one could choose from many different functions to implement this prior; however, in this work we focus on using a Gaussian mixture model (GMM) building on our previous use of GMM grasp priors in~\cite{lu2019grasp}.

As a third prior, we propose an object conditional prior as shown in Eq.~\ref{eq:inf_MDN}.
\begin{equation}
  p(\boldsymbol{\theta} | Y=1, z, \bm{W}, \bm\Phi) \propto p(Y=1 | \boldsymbol{\theta}, z, \bm{W}) p(\boldsymbol{\theta} | z, \bm\Phi) \label{eq:inf_MDN}
\end{equation}
Here, the preferred grasp configurations are conditioned on the observed sensory information. For both the GMM of Eq.~\ref{eq:inf_GMM} and the object conditional prior of Eq.~\ref{eq:inf_MDN} we examine data-driven priors to encode knowledge of what data was observed by the learner during training. This prior thus prefers grasps similar to those seen during training. Such priors can be viewed as an approximation of the epistemic uncertainty of the learned classifier~\cite{Kendall2017}, encoding the belief that the classifier's confidence should decrease for grasp or objects far from those observed during training.  We define the details of the object-conditional prior as a deep neural network in Sec.~\ref{subsec:voxel_prior}.
We perform experiments comparing grasp inference using variants of these three priors in Section~\ref{sec:exp}.

We solve the inference problem for all three grasp models in the log-probability space and regularize the log-prior with a multiplicative gain of $0.5$ to prevent the prior dominating the inference. We use the popular L-BFGS optimization algorithm with bound constraints~\cite{byrd1995limited}~\cite{zhu1997algorithm} to efficiently solve the inference problem. We use the scikit-learn\footnote{http://scikit-learn.org/stable/index.html} library to perform the optimization. We initialize the inference by randomly sampling from the learned priors. We initialize the uniform prior using a heuristic described in Sec.~\ref{subsec:data_collection} as previously done~\cite{lu2017grasp}.


\section{Voxel-Based Deep Networks for Multi-fingered Grasp Learning}
\label{sec:grasp_learning}
In this section we present the design of our voxel-based neural network classifier for predicting grasp success on multi-fingered hands. We also describe the structure of our voxel-based mixture density network (MDN) that models the object-conditional prior probability over grasp configurations. We then describe the data collection and training of these networks, before showing that our voxel-based classifier achieves better offline testing performance for grasp success probability prediction compared to our previous RGB-D-based neural network classifier in Sec.~\ref{subsec:offline-validation}.

\begin{figure*}
	\centering
	\begin{subfigure}[]{0.65\textwidth}
		\includegraphics[width=\textwidth]{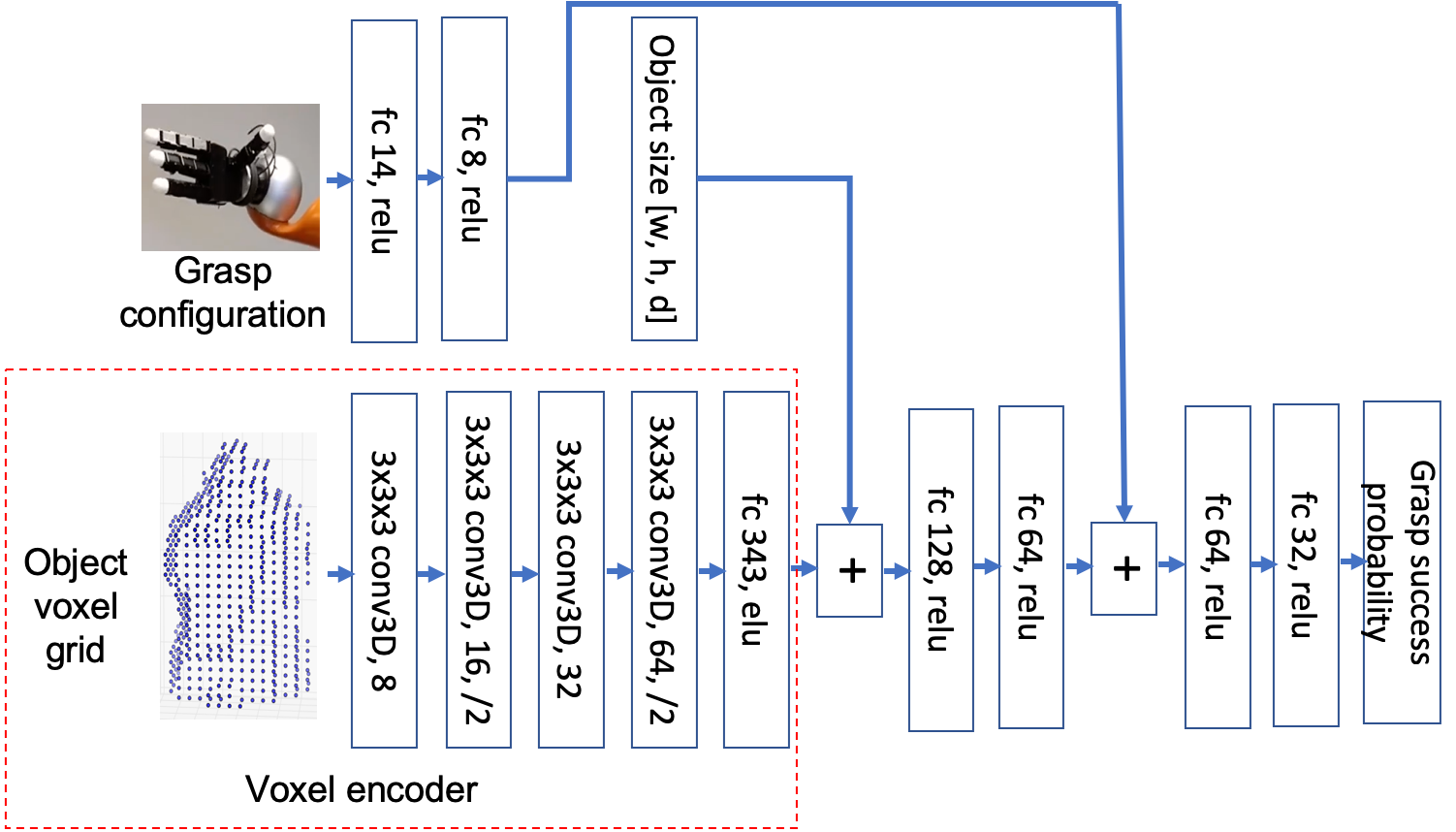}
		\caption{The architecture of our voxel-config-net for grasp success probability prediction.}
		\label{fig:voxel-config-net}
	\end{subfigure}
	\begin{subfigure}[]{0.65\textwidth}
	\includegraphics[width=\textwidth]{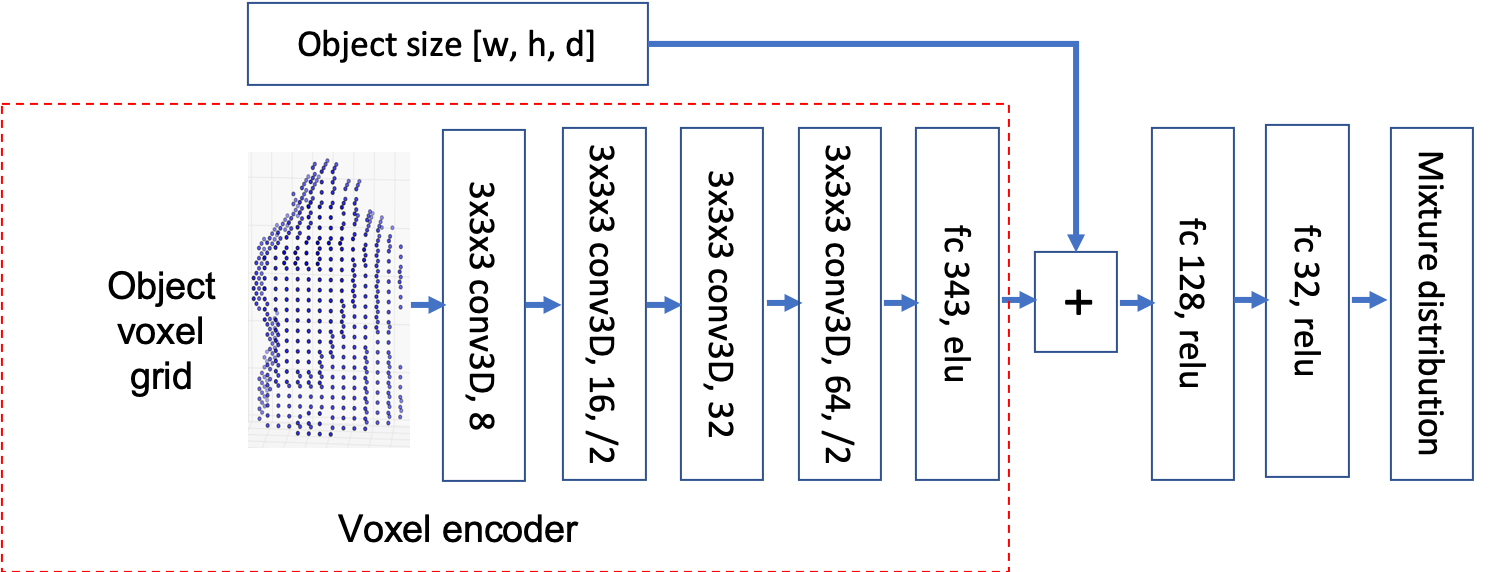}
	\caption{The architecture of our MDN modeling the grasp conditional prior.}
	\label{fig:mdn_arch}
\end{subfigure}
	\caption{The voxel-config-net and MDN architectures. Bottom left visualizes the voxel-grid for the ``mustard bottle'' object. All convolutional layers use $3 \times 3 \times 3$ 3D convolutional filters with exponential linear unit (ELU) activations. We annotate the number of filters and the stride ($/2$ means a stride of 2) for convolutional layers. We annotate the number of neurons and the activation function for fully connected layers.}
	\label{fig:voxel-nets}
\end{figure*}

\subsection{Voxel-Based Grasp Likelihood Classification}
\label{subsec:voxel_classifier}
Figure~\ref{fig:voxel-config-net} shows the architecture of our grasp success prediction network. Our voxel-based classifier takes three inputs: a $32 \times 32 \times 32$ object voxel-grid, a vector defining the width, height, and depth of the voxel-grid in the 3D scene, and a $14$ dimensional vector encoding the grasp preshape configuration, which we define in more detail in Sec.~\ref{subsec:data_collection}. We note this preshape configuration could be replaced with other grasp representations in a straightforward manner. The network processes the object voxel-grid with a sub-network composed of four 3D convolutional layers and one fully-connected layer, which we name the ``voxel encoder''. We pre-train this voxel encoder on a 3D object reconstruction task described in Sec.~\ref{subsec:train_eval}.

We concatenate the voxel features processed by the voxel encoder with the object size vector, and pass the concatenated features through two fully-connected layers to generate the final object feature representation.
The grasp configuration input is processed by two fully-connected layers to generate the grasp configuration features. Then we concatenate the grasp configuration features and the final object features, and pass them through two fully-connected layers followed by a sigmoid output layer to generate the grasp success probability. We apply batch normalization for all convolutional and fully-connected layers except the output layer. We train our voxel-based classifier using the cross entropy loss.

In order to generate the voxel-grid we first segment the object from the 3D point cloud by fitting a plane to the table using RANSAC~\cite{Fischler1981,hermans-icra2013} and extracting the points above the table. We then estimate the first and second principle axes of the segmented object to create a right-handed object reference frame aligned relative to the world frame.
We compute the object size along the three coordinates of the object reference frame to construct the object size vector. 
We then generate a $32 \times 32 \times 32$ voxel grid oriented about this reference frame. We define the center of the voxel grid to be the centroid of the points in the object segmentation. More details of the point cloud voxelization can be seen from Section~\ref{subsec:train_eval}.

\subsection{Voxel-Based Grasp Prior Networks}
\label{subsec:voxel_prior}
In order to model the grasp configuration distribution based on the geometry of the object of interest, we construct a mixture-density network (MDN) as our object conditional prior. Given its input an MDN predicts the parameters (means, covariance and mixing weights) of a Gaussian mixture model as output. Our MDN takes the object voxel-grid and the object size vector as inputs and predicts the parameters of a GMM modeling a probability distribution over grasp configurations. Thus the MDN learns to model the conditional probability distribution \(p(\boldsymbol{\theta} | z, \bm\Phi)\) where \(\bm\Phi\) define the learned weights of the MDN. We train the MDN over all grasp attempts from the training set, meaning the learned distribution models the probability that the specific grasp \(\boldsymbol{\theta}\) being evaluated was observed at training given the current object.

The MDN generates its object feature representation using the same sub-network structure as voxel-based classifier. The MDN then passes the final object feature representation through two fully-connected layers with ReLU activations. These two fully-connected layers have $128$ and $32$ neurons respectively. Finally, the fully-connected output layer predicts the weights, mean, and diagonal covariance of the mixture distribution over grasp configuration. Figure~\ref{fig:mdn_arch} visualizes the architecture of our MDN. We apply batch normalization for all layers of the MDN except the output layer. We train our voxel-based MDN using the negative log likelihood loss.

\begin{figure*}[]
	\centering
	\includegraphics [width=0.75\textwidth] {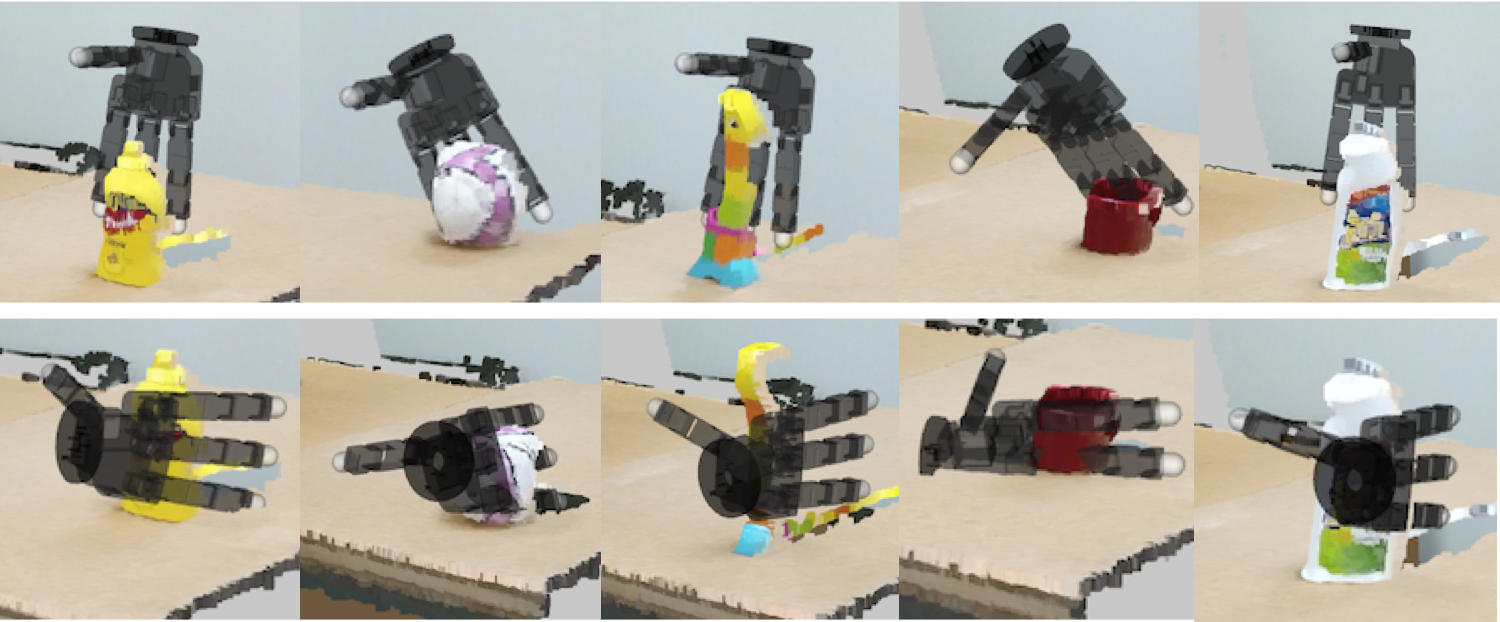}
	\caption{The mean visualizations of these two MDN mixture components for five different objects. Top row shows mean of the overhead component and botton row shows mean of the side component.}
	\label{fig:mdn_means}
\end{figure*}

We visualize the mean grasp configuration of each mixture component predicted by the MDN for several different objects in Fig.~\ref{fig:mdn_means}.


\subsection{Grasp Data Collection}
\label{subsec:data_collection}
We conduct all training and experiments using the four-fingered, 16 DOF Allegro hand mounted on a Kuka LBR4 7 DOF arm. We use a Kinect2 camera to generate the point cloud of the object on the table.
We collected simulated grasp data using our robot hand-arm setup inside the Gazebo simulator with the DART physics engine\footnote{\url{https://dartsim.github.io/}}. We use the built-in Gazebo Kinect camera to generate point clouds simulating a Kinect2 RGB-D camera we use in real-world experiments. All data and software used in this paper are available online\footnote{\url{https://robot-learning.cs.utah.edu/project/grasp_voxel_inference}}.

We collected training data using a heuristic, geometry-based grasp planner adapted from our previous planner in~\cite{lu2017grasp}, which is quite similar to the geometric primitive planner of~\cite{miller2003automatic} for boxes or cylinders. We collected both multi-fingered side and overhead grasps. For side grasps we have the thumb pointing towards the top as in~\cite{lu2017grasp}. In~\cite{lu2017grasp}, overhead grasps were randomly selected to align with either the major or minor axis of the bounding box top face, which fails to generate overhead grasps robustly.
In this article, we improve the overhead heuristic planner by aligning the thumb to point in the direction of the mean vector of the major and minor axes of the bounding box top face with the palm parallel to the top face. This modification boosts the overhead grasp success rates in data collection.

We generate a preshape by randomly sampling joint angles for the first two joints of all fingers within a reasonable range, fixing the last two joints of each finger to be zero.
There are 14 parameters for the Allegro hand preshape, 6 for the palm pose and 8 relating to the first 2 joint angles of each finger proximal to the palm. Given a desired pose and preshape we use the RRT-connect motion planner in MoveIt! to plan a path for the arm. We execute all feasible plans moving the robot to the sampled preshape.

After moving the hand to the desired preshape, the robot runs a grasp controller to close the hand. The grasp controller closes the fingers at a constant velocity stopping each finger independently when contact is detected by the measured joint velocities being close to zero.
The grasp controller closes the second and third joints of the non-thumb fingers and the two distal joints of the thumb. Note the proximal joint of all non-thumb fingers rotates the finger about its major axis causing it to change the direction of closing. As such we maintain the angle provided by the grasp planner for these joints.
\begin{table}[]
	\begin{center}
		\begin{tabular}{|l|l|l|l|l|}
			\hline
			\multirow{2}{*}{Grasp Type} & \multicolumn{2}{l|}{Training} & \multicolumn{2}{l|}{Offline Testing} \\ \cline{2-5}
			& Success       & Failure       & Success       & Failure      \\ \hline
			Side       &    1559             &     2616          &        339      &       515                 \\ \hline
			Overhead   &   749              &      4064         &       157       &      810                 \\ \hline
		\end{tabular}
		\caption{Number of successful and failure grasps of side and overhead grasps in the simulation-based training and offline testing sets generated using the heuristic planner.}
		\label{tab:data_collection}
	\end{center}
\end{table}

Upon closing, the robot attempts to lift the object to a height of $15$cm. If the robot succeeds in reaching this height without the object falling, the simulator automatically labels the grasp as successful. We collected $10,809$ grasp attempts in total of which $2,804$ resulted in successful grasps. The dataset covers more than $100$ objects of the Bigbird~\cite{singh2014bigbird} dataset. This dataset contains more than 7 times the number of grasp attempts as in the dataset from our previous work~\cite{lu2017grasp}.
This dataset contains $1,898$ successful side grasps and $906$ successful overhead grasps.
We use $8,988$ grasps for training of our grasp model and $1,821$ grasps for testing. Table~\ref{tab:data_collection} shows the number of successful and failed grasp attempts of side and overhead grasps separately in our training and offline testing sets.

\subsection{Grasp Model Training}
\label{subsec:train_eval}
We train our networks using the Adam optimizer with mini-batches of size $64$ for $90$ epochs. The learning rate starts at $0.001$ and decreases by $10\times$ every $30$ epochs. The MDN is trained with the same specifications.
The training of both models take less than $25$ minutes on a computer with an Intel-i74790k processor, 64GB RAM, and an Nvidia GeForce GTX 970 graphics card.
We implement all our deep network models in TensorFlow. We fit the GMM prior parameters using the EM algorithm over all grasp configurations attempted in the training set.

We pre-train the voxel encoder for our classifier and MDN on a voxel-based 3D object reconstruction task. We freeze the voxel encoder parameters to the values learned on this reconstruction task during grasp training of the classifier and MDN. We found freezing the encoder parameters achieves better testing performance than fine-tuning the encoder parameters for both models.

Our voxel reconstruction autoencoder has the same output and decoder structure as the reconstruction variational autoencoder in~\cite{brock2016generative}, except we drop the variational constraint, \emph{i.e.} we do not predict a variance associated with the output of the encoder. We found this achieves better reconstruction results than the variational autoencoder in~\cite{brock2016generative} on our dataset.

To train our voxel reconstruction autoencoder, we synthetically render 590 meshes from the Grasp Database~\cite{Kappler2015} at 200 random orientations each, adding noise to the depth images to reflect sensor noise. We backproject these points into a 3D point cloud, which we voxelize to a $26 \times 26 \times 26$ voxel-grid, centered in a $32 \times 32 \times 32$ total voxel-grid.  We scale the bounding box extracted from the segmented object pointcloud to have $26 \times 26 \times 26$ voxels. We are here concerned only with capturing object shape information, as object size and pose are handled by the grasp network independently. As such, we learn the mapping from the partial point cloud voxelization to a \textit{centered} and independently scaled full mesh voxelization. We train with a sigmoid cross entropy loss on the true and predicted reconstructed voxel representation. The voxel reconstruction autoencoder is trained with the momentum optimizer with mini-batches of size $64$ for $100$ epochs. We use a starting learning rate of $0.001$ and decrease the rate by $10\times$ after $20$ and $80$ epochs and set momentum to $0.9$. Our network trains in about $4$ hours on a single Nvidia Tesla V100 graphics card. Our pretrained, voxel reconstruction network achieved 95.56\% accuracy and an F1-score of 0.5166 on a test set comprised of left out models from the Grasp Database~\cite{Kappler2015} and all models from the YCB dataset~\cite{calli2015benchmarking}.

\subsection{Offline Grasp Learning Validation}
\label{subsec:offline-validation}
We validated the performance of our voxel-based grasp success classifier on an offline prediction task using the held-out test set collected in simulation. We show the prediction accuracy and F1 score in Table~\ref{tab:model_eval}. We compare our approach (voxel-config-net) with our previous RGB-D-based classifier from~\cite{lu2017grasp} (rgbd-config-net). We retrain the RGB-D network on the training data of this article with the same specifications as above, except we use mini-batches of size $8$. The RGB-D network takes more than $24$ hours to finish $90$ epochs of training.
\begin{table}[h!]
	\begin{center}
		\begin{tabular}{|l|l|l|l|l|l|}
			\hline
			\multirow{2}{*}{Grasp Type} & \multicolumn{2}{l|}{voxel-config-net} & \multicolumn{2}{l|}{RGB-D-config-net} & MDN   \\ \cline{2-6}
			& Accuracy            & F1              & Accuracy           & F1              & Loss  \\ \hline
			Both                         & 0.786               & 0.592           & 0.573              & 0.476           & -10.73 \\ \hline
			Side                        & 0.724               & 0.649           & 0.661              & 0.401           & -14.71 \\ \hline
			Overhead                    & 0.842               & 0.456           & 0.474              & 0.52            & -7.2 \\ \hline
		\end{tabular}
		\caption{Grasp model evaluation on testing set for all, side and overhead grasps. We show the accuracy and F1 score of voxel-config-net and RGB-D-config-net. We also show the negative log likelihood loss of MDN.}
		\label{tab:model_eval}
	\end{center}
\end{table}

We see our voxel-based network significantly outperforms the classifier using the RGB-D object representation on this task in terms of both F1 and accuracy. This result holds for both side and overhead grasps. We believe this implies the voxel-based approach encodes the object geometry better for grasping than using an RGB-D image directly.

We treat predictions with grasp probability above $0.5$ as positive for the voxel-based approach. As done in~\cite{lu2017grasp} we threshold the RGB-D network predictions with a value of $0.4$, as setting the threshold to 0.5 predicts all grasps as failures. This necessary modification highlights another advantage of our novel voxel-based classifier, namely that the predicted grasp probabilities better reflect the true success probabilities making them more useful for planning as inference.

For completeness, we also show the MDN negative log likelihood loss on the testing set, where smaller MDN loss reflects higher conditional probability density. 




\section{Robotic Grasp Inference Experiments}
\label{sec:exp}
\begin{figure}[t!]
	\centering
	\includegraphics[ width=0.8 \linewidth]{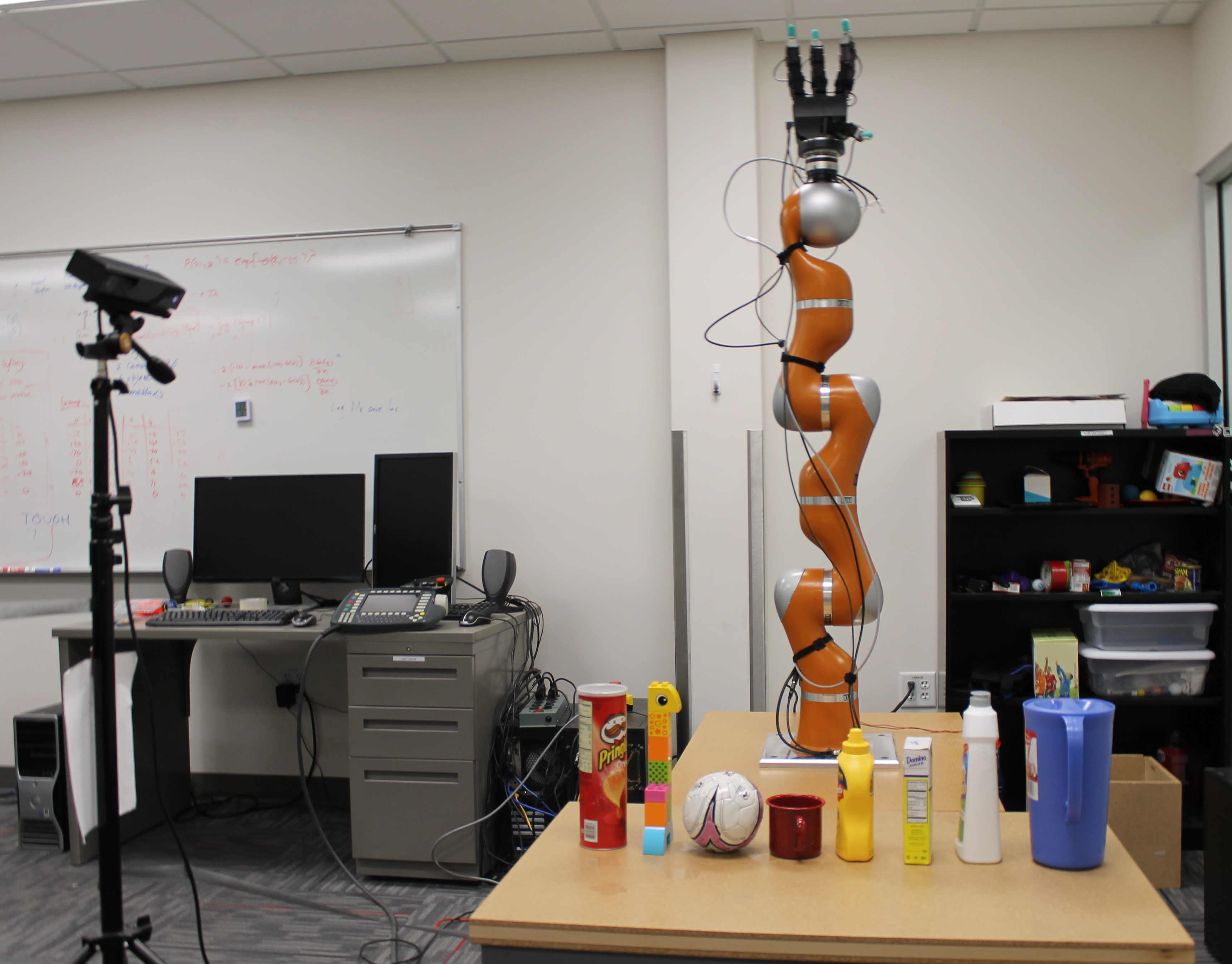}
	\caption{Experimental setup with objects used for experiments. From left to right objects are ``pringles'', ``Lego'', ``soccer ball'', ``mug'', ``mustard bottle'', ``sugar box'', ``soft scrub'', and ``pitcher.'' Objects range in size from \(8 \times 9 \times 11\)cm (mug) to \(13 \times 17 \times 24\)cm (pitcher).}
	\label{fig:exp_setup}
      \end{figure}

We evaluate grasp planning as inference on the physical robot system using our learned voxel-based grasp classifier with all three grasp configuration prior probability models described in Section~\ref{sec:grasp_inference} (i.e. uniform, GMM, and object-conditional). We perform experiments on 8 YCB~\cite{calli2015benchmarking} objects covering different textures, shapes, and sizes. We show the experimental setup and objects used in Figure~\ref{fig:exp_setup}.  All experimental objects are unseen in training except for ``Pringles''. We attempted both overhead and side grasps at $5$ different poses per object, for a total of $80$ grasp attempts per method. We use the same set of locations across different methods, but each object has its own set of random poses. In total, we performed $240$ grasp attempts for $3$ different methods across $8$ objects in this article.

Our evaluation protocol on the physical robot mirrors that used in simulation. Namely, we label a grasp attempt that successfully lifts the object to a height of $0.15$m without dropping it as successful. We use the same motion planner and grasp controller as in data collection to plan paths for the arm and close the hand. If the motion planner fails to generate a plan for a grasp due to either inverse-kinematics (IK) or collision avoidance, we generate a new grasp using the same grasp planner with a different initialization. If the grasp planner could not generate a grasp with a motion plan in $5$ attempts, we treat the grasp attempt as a failure case.

The MDN object conditional prior and GMM prior both have two mixture components. For both learned prior models, we found the grasp configuration mean of one mixture component is a side grasp (we term this component the ``side grasp component'') and the grasp configuration mean of the other component is an overhead grasp (we term this component the ``overhead grasp component''). We randomly sample a grasp configuration from the side grasp component to initialize the grasp inference to generate a side grasp. Similarly, we initialize the grasp inference with a random sample from the overhead grasp component to plan an overhead grasp. We initialize the grasp inference of the uniform prior with side and overhead grasps generated by the heuristic geometry-based planner used for data collection to plan side and overhead grasps respectively.

\begin{figure}[]
	\centering
	\begin{adjustbox}{minipage={\columnwidth}}
		\begin{tikzpicture}
		\begin{axis}[
		ybar, ymin=-1, ymax=100,
		ylabel={Success Rate (\%)},
		axis lines*=left,
		symbolic x coords={Pringles, Pitcher, Lego, Mustard, Mug, Soft scrub,  Sugar box, Soccer ball, All},
		xtick=data,
		ticklabel style = {font=\scriptsize, rotate=45},
		legend style={font=\scriptsize, draw=none, fill=none},
		y label style={at={(axis description cs:0.02,0.5)},anchor=north},
		bar width = 4pt, height=4.5 cm, width=\linewidth,
		legend style={area legend, at={(1,1.20)}, anchor=north east, legend columns=4, },
		legend image code/.code={%
			\draw[#1] (0cm,-0.1cm) rectangle (2mm,1mm);},
		]
		\addplot [fill=purple, postaction={pattern=north east lines}] coordinates {
			(Pringles, 100) (Pitcher, 100) (Lego, 100) (Mustard, 100) (Mug, 0) (Soft scrub, 100) (Sugar box, 80) (Soccer ball, 20)
			(All, 75)};

		\addplot [fill=yellow, postaction={pattern=horizontal lines}] coordinates {
			(Pringles, 100) (Pitcher, 80) (Lego, 60) (Mustard, 80) (Mug, 0) (Soft scrub, 60) (Sugar box, 80) (Soccer ball, 0)
			(All, 58)};

		\addplot [fill=orange] coordinates {
			(Pringles, 0) (Pitcher, 0) (Lego, 0) (Mustard, 0) (Mug, 0) (Soft scrub, 0) (Sugar box, 20) (Soccer ball, 0)
			(All, 3)};
		\legend{Voxel-MDN-prior, Voxel-GMM-prior, Voxel-uniform-prior}
		\end{axis}
		\end{tikzpicture}
	\end{adjustbox}
	\caption{Multi-fingered side grasping success rates of $3$ different prior distribution methods on the real robot. ``Pringles'' was seen in training, other $7$ objects are previously unseen.}
	\label{fig:side_voxel_results}
\end{figure}
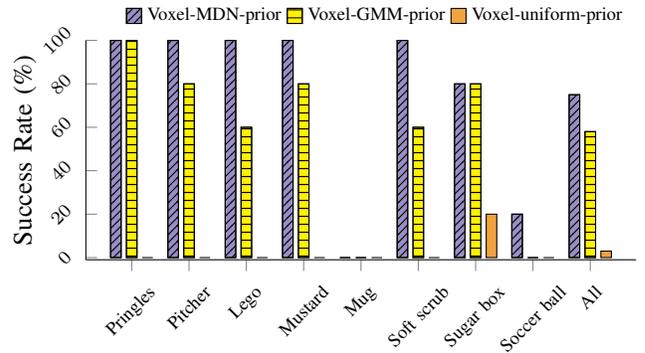

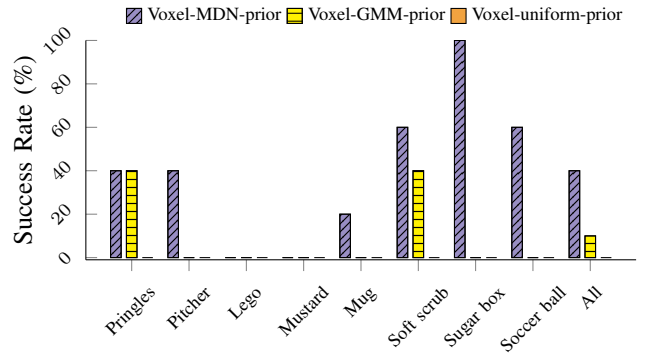
\begin{figure}[]
	\centering
	\begin{adjustbox}{minipage={\columnwidth}}
		\begin{tikzpicture}
		\begin{axis}[
		ybar, ymin=-1, ymax=100,
		ylabel={Success Rate (\%)},
		axis lines*=left,
		symbolic x coords={Pringles, Pitcher, Lego, Mustard, Mug, Soft scrub,  Sugar box, Soccer ball,  All},
		xtick=data,
		ticklabel style = {font=\scriptsize, rotate=45},
		legend style={font=\scriptsize, draw=none, fill=none},
		y label style={at={(axis description cs:0.02,0.5)},anchor=north},
		bar width = 4pt, height=4.5 cm, width=\linewidth,
		legend style={area legend, at={(1,1.20)}, anchor=north east, legend columns=4, },
		legend image code/.code={%
			\draw[#1] (0cm,-0.1cm) rectangle (2mm,1mm);},
		]
		\addplot [fill=purple, postaction={pattern=north east lines}] coordinates {
			(Pringles, 40) (Pitcher, 40) (Lego, 0) (Mustard, 0) (Mug, 20) (Soft scrub, 60) (Sugar box, 100) (Soccer ball, 60)
			(All, 40)};

		\addplot [fill=yellow, postaction={pattern=horizontal lines}] coordinates {
			(Pringles, 40) (Pitcher, 0) (Lego, 0) (Mustard, 0) (Mug, 0) (Soft scrub, 40) (Sugar box, 0) (Soccer ball, 0)
			(All, 10)};

		\addplot [fill=orange] coordinates {
			(Pringles, 0) (Pitcher, 0) (Lego, 0) (Mustard, 0) (Mug, 0) (Soft scrub, 0) (Sugar box, 0) (Soccer ball, 0)
			(All, 0)};
		\legend{Voxel-MDN-prior, Voxel-GMM-prior, Voxel-uniform-prior}
		\end{axis}
		\end{tikzpicture}
	\end{adjustbox}
	\caption{Multi-fingered overhead grasping success rates of $3$ different methods on the real robot. ``Pringles'' was seen in training, other $7$ objects are previously unseen.}
	\label{fig:overhead_voxel_results}
\end{figure}

The side grasp success rates for all three methods are summarized in Figure~\ref{fig:side_voxel_results}. Figure~\ref{fig:overhead_voxel_results} shows the overhead grasp success rates for all three methods. It takes $5-10$ seconds for each method to generate a grasp.
The grasp planners using the MDN prior, the GMM prior, and the uniform prior achieve success rates of $75\%$, $58\%$, and $3\%$ respectively on side grasps for the $8$ objects. The grasp planner is not able to plan any successful side grasps for the object mug. Since the mug object is relatively short, the motion planner could not find paths which would not collide with the table for side grasps for all three methods.
The grasp planners using the MDN prior, GMM prior, and uniform prior achieve success rates of $40\%$, $10\%$, and $0\%$ respectively on overhead grasps the $8$ objects. 
The grasp planner never generates overhead grasps successfully for the objects mustard and lego.
Mustard and lego have relatively smaller contact areas available for overhead grasps and our grasp controller would push them away when closing the hand as it had no feedback from vision or haptic sensors to know the object was moving.

The grasp planner using the MDN prior achieves higher success rates than the two other methods for both side and overhead grasps, which demonstrates the benefit of modeling the grasp prior conditionally on the observed object when performing inference.
The uniform prior achieved the lowest success rates for both side and overhead grasps. This shows that data-driven priors, even if not conditioned on the observed object, outperform using a heuristic planner for initialization and locally constraining the grasp configuration, as these heuristics can not reliably generate successful grasps.

\begin{figure*}[h]
	\centering
	\includegraphics[ width=0.9\linewidth]{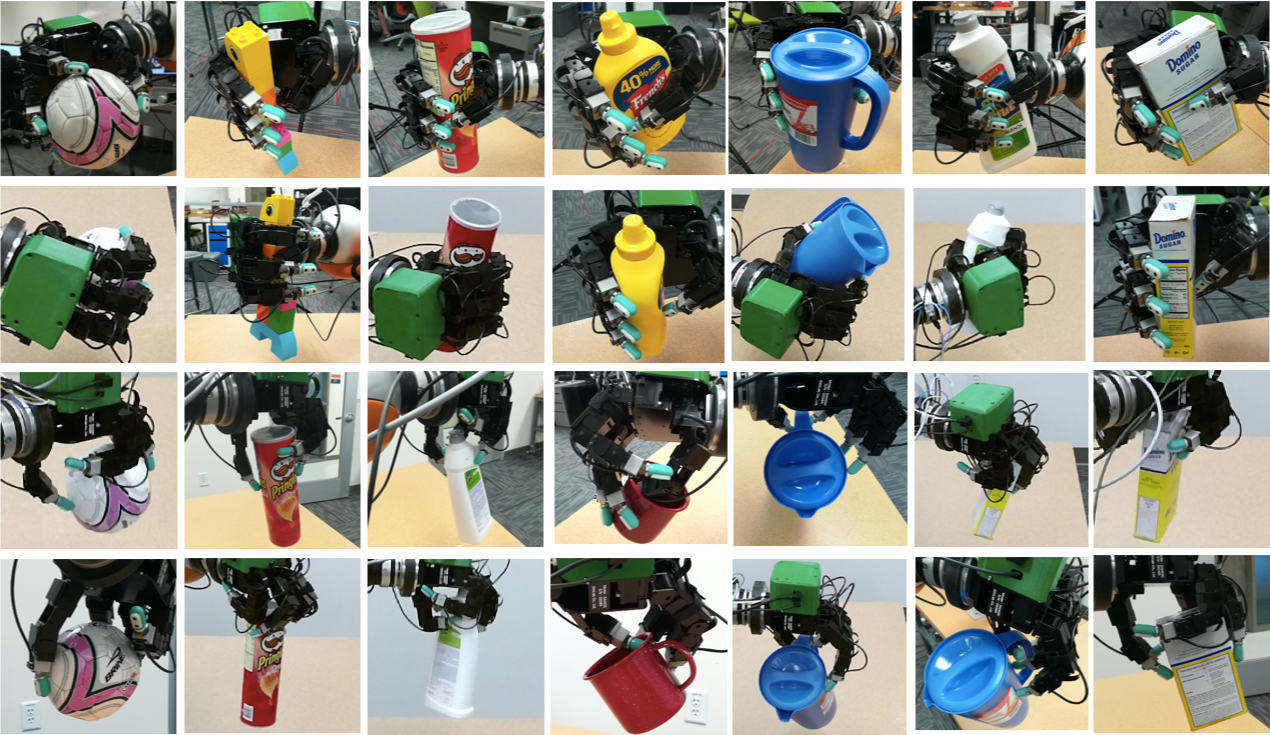}
	\caption{Examples of successful grasps generated by grasp inference using our voxel-based classifier with the MDN object-conditional prior. The top two rows are side grasps. Bottom two rows are overhead grasps.}
	\label{fig:voxel_grasp_examples}
\end{figure*}

In Figure~\ref{fig:voxel_grasp_examples}, we show example grasps for different objects generated by our inference approach with the voxel-based classifier and MDN prior. Grasps in the top two rows are side grasps, which provide strong stability. The bottom row shows overhead grasps which provide access to objects in clutter and often provide improved dexterity between the robot and object.

\section{Discussion \& Conclusions}
\label{sec:conclusions}
In this article we presented a novel approach for multi-fingered grasp planning formulated as probabilistic inference in learned deep neural networks. We proposed a voxel-based 3D convolutional neural network to predict the probability of grasp success as a function of both the object voxel-grid and grasp configuration. We show that the our novel, voxel-based grasp success classifier for multi-fingered grasping outperforms our previous RGB-D image patch based neural network presented in~\cite{lu2017grasp} in terms of both predictive accuracy and training time.

In our previous work, we showed that planning as inference with the RGB-D structure outperforms several alternatives for grasp planning, such as sampling and regression. As our results here indicate improvement over the RGB-D network structure, we can infer that our planning similarly outperforms alternative multi-fingered grasp planning approaches, while still running fast enough for use in a deployed robotic system. Additionally, our CNN classification approach to learning grasp success allows for more data-efficient learning compared to directly predicting grasps using regression. In the regression-based formulation the neural network takes the visual information (e.g. RGB-D or voxel-grid) as input and directly predicts a grasp configuration as output. These regression models can learn only from successful grasps, while our success classifier learns from both successful and failed grasp attempts alike.

Using our learned voxel-based classifier we examined the role of different prior probability distributions over grasp configuration in the planning process. Our real-robot grasp experiments for the three different prior models defined in Section~\ref{sec:grasp_inference} show that using a learned object conditional prior over grasp configurations benefits grasp inference when combined with the learned grasp success prediction network. This learned mixture density network benefits from the same voxel-encoder to represent the observed object of interest.
Furthermore the data-driven Gaussian mixture model also provides benefits over the bounded uniform prior. This provides further evidence that learned priors provide better planning performance compared with heuristic, weak priors.

Learned priors provide additional benefits for multi-fingered grasp planning. First, they enable the robot to directly sample initial configurations for use in the resulting optimization problem. In our previous work, we showed how randomly generating numerous samples from the uniform prior and selecting the one with highest predicted success fails to reliably generate successful grasps. This demonstrates the benefit of the learned prior, which focuses the search space to promising configurations, something the uniform prior cannot provide.
To overcome this problem previously we relied on an external grasp planner to initialize the optimization.

The second additional benefit of the learned prior comes from it removing the need for an external planner or heuristic for initialization. In addition to generally being computationally more efficient, removal of this external planner reduces the bias present in human-designed planners which limit the space of grasps under consideration. Model-based planners tend to prefer only a single class of grasps such as power or precision and not both~\cite{lu2019grasp}. By leveraging data-driven priors the robot is not restricted to grasps similar to those provided by the planner, but instead can leverage any grasp its learned model predicts will be successful. Indeed, our resulting grasp planner reliably generates successful side and overhead grasps on the real robot across several different objects used for testing.

We can directly attribute this ability to generate both overhead and side grasps to the new dataset we generated for this article, that contains substantially more successful overhead grasps than our previous dataset from~\cite{lu2017grasp}.
However, this improvement highlights that we have simply shifted the burden of the external planner from inference initialization to grasp exploration in generating training data. While we overcome the bias of the planner somewhat by adding random perturbations to the output of our heuristic planner, we still limit the space of grasps explored during training.

In order to overcome this issue in the future, we wish to explore active learning where the robot will select what grasps to attempt for learning based on the previous attempted grasps and the currently learned grasp model. This should both improve the data efficiency of our learning algorithm while also learning a wider variety of grasps. However, new issues arise of how to correctly update the learned model in an online fashion as the standard IID data assumption used for batch neural network training will no longer hold.

A more obvious shortcoming of our mixture density network prior stems from the need to explicitly select the number of mixture components in the model. An open question remains as to how we can expand the capacity of the mixture network to encode a higher variety of grasps as the robot collects more data for training.

As we noted in Section~\ref{sec:grasp_inference} our learned priors can be viewed as an approximation of the model (i.e. epistemic~\cite{Kendall2017}) uncertainty of the learned classifier. In future work we wish to compare prior learning with explicitly learning priors over the neural network weights \(w\), which would hopefully provide better-calibrated predictions of the probability of grasp success. Accurate models of the probability of grasp success would enable more reliable task-level planning, where the robot could reason over the probability of a sequence of events producing the desired outcome under uncertainty of the manipulated object's shape and physical properties. However it is unclear how one could use such Bayesian neural networks to efficiently perform MAP inference for grasp planning, as a single evaluation of the neural network uncertainty typically requires several forward pass evaluation of the neural network model~\cite{Kendall2017}.

A final weakness of our results as presented stems from our planner achieving better performance in attempted side grasps than overhead grasps. This presents significant issues in attempting to perform grasping in clutter or of low-profile objects where the hand must be close to the table to grasp the object. Indeed this problem arose in attempting to grasp the mug in this article. We think learning or designing a more complex feedback controller for overhead grasps using tactile feedback would boost the overhead grasp performance, especially for these objects with less contact areas on the top.

In conclusion our article shows that we can improve grasp planning as MAP inference by incorporating three particular benefits. First, using a voxel-based object representation instead of an RGB-D improves learning performance. Second, learning mixture density network priors improves over uniform or object-independent learned priors. Three, unsurprisingly more data representing grasps of increased variability improves grasp planning.

Nevertheless several issues and open questions still remain with our planning framework as present. Clearly learning-based approaches are becoming more and more prevalent if not the norm for manipulation. We hope the manipulation community takes hold of these questions and finds more to further our understanding of grasp planning as probabilistic inference.

\bibliographystyle{IEEEtran}
\bibliography{grasp_ref}  

\begin{thebibliography}{10}
\providecommand{\url}[1]{#1}
\csname url@samestyle\endcsname
\providecommand{\newblock}{\relax}
\providecommand{\bibinfo}[2]{#2}
\providecommand{\BIBentrySTDinterwordspacing}{\spaceskip=0pt\relax}
\providecommand{\BIBentryALTinterwordstretchfactor}{4}
\providecommand{\BIBentryALTinterwordspacing}{\spaceskip=\fontdimen2\font plus
\BIBentryALTinterwordstretchfactor\fontdimen3\font minus
  \fontdimen4\font\relax}
\providecommand{\BIBforeignlanguage}[2]{{%
\expandafter\ifx\csname l@#1\endcsname\relax
\typeout{** WARNING: IEEEtran.bst: No hyphenation pattern has been}%
\typeout{** loaded for the language `#1'. Using the pattern for}%
\typeout{** the default language instead.}%
\else
\language=\csname l@#1\endcsname
\fi
#2}}
\providecommand{\BIBdecl}{\relax}
\BIBdecl

\bibitem{saxena2006learning}
A.~Saxena, J.~Driemeyer, and A.~Y. Ng, ``{Robotic Grasping of Novel Objects
  using Vision},'' \emph{The International Journal of Robotics Research
  (IJRR)}, vol.~27, no.~2, pp. 157--173, 2008.

\bibitem{Saxena-aaai2008}
A.~Saxena, L.~L.~S. Wong, and A.~Y. Ng, ``{Learning Grasp Strategies with
  Partial Shape Information},'' in \emph{AAAI National Conf. on Artificial
  Intelligence}, 2008, pp. 1491--1494.

\bibitem{lenz2015deep}
I.~Lenz, H.~Lee, and A.~Saxena, ``{Deep Learning for Detecting Robotic
  Grasps},'' \emph{The International Journal of Robotics Research (IJRR)},
  vol.~34, no. 4-5, pp. 705--724, 2015.

\bibitem{pinto2016supersizing}
L.~Pinto and A.~Gupta, ``{Supersizing Self-supervision: Learning to Grasp from
  50K Tries and 700 Robot Hours},'' in \emph{IEEE Intl. Conf. on Robotics and
  Automation (ICRA)}, 2016, pp. 3406--3413.

\bibitem{Kopicki2016}
M.~Kopicki, R.~Detry, M.~Adjigble, R.~Stolkin, A.~Leonardis, and J.~L. Wyatt,
  ``{One-Shot Learning and Generation of Dexterous Grasps for Novel Objects},''
  \emph{The International Journal of Robotics Research (IJRR)}, vol.~35, no.~8,
  pp. 959--976, 2016.

\bibitem{mousavian20196}
A.~Mousavian, C.~Eppner, and D.~Fox, ``6-dof graspnet: Variational grasp
  generation for object manipulation,'' \emph{arXiv preprint arXiv:1905.10520},
  2019.

\bibitem{wu2019pixel}
B.~Wu, I.~Akinola, and P.~K. Allen, ``Pixel-attentive policy gradient for
  multi-fingered grasping in cluttered scenes,'' \emph{arXiv preprint
  arXiv:1903.03227}, 2019.

\bibitem{liu2019generating}
M.~Liu, Z.~Pan, K.~Xu, K.~Ganguly, and D.~Manocha, ``Generating grasp poses for
  a high-dof gripper using neural networks,'' \emph{arXiv preprint
  arXiv:1903.00425}, 2019.

\bibitem{sahbani2012overview}
A.~Sahbani, S.~El-Khoury, and P.~Bidaud, ``{An Overview of 3D Object Grasp
  Synthesis Algorithms},'' \emph{Robotics and Autonomous Systems}, vol.~60,
  no.~3, pp. 326--336, 2012.

\bibitem{bohg2014data}
J.~Bohg, A.~Morales, T.~Asfour, and D.~Kragic, ``{Data-Driven Grasp Synthesis-a
  Survey},'' \emph{IEEE Transactions on Robotics}, vol.~30, no.~2, pp.
  289--309, 2014.

\bibitem{ciocarlie2007dexterous}
M.~Ciocarlie, C.~Goldfeder, and P.~Allen, ``{Dimensionality Reduction for
  Hand-Independent Dexterous Robotic Grasping},'' in \emph{IEEE/RSJ Intl. Conf.
  on Intelligent Robots and Systems (IROS)}, 2007, pp. 3270--3275.

\bibitem{dragiev2011gaussian}
S.~Dragiev, M.~Toussaint, and M.~Gienger, ``{Gaussian Process Implicit Surfaces
  for Shape Estimation and Grasping},'' in \emph{IEEE Intl. Conf. on Robotics
  and Automation (ICRA)}, 2011, pp. 2845--2850.

\bibitem{Grupen1991}
R.~A. Grupen, ``{Planning Grasp Strategies for Multifingered Robot Hands},'' in
  \emph{IEEE Intl. Conf. on Robotics and Automation (ICRA)}, 1991, pp.
  646--651.

\bibitem{Murray1994}
R.~M. Murray, Z.~Li, and S.~S. Sastry, \emph{{A Mathematical Introduction to
  Robotic Manipulation}}.\hskip 1em plus 0.5em minus 0.4em\relax CRC Press,
  1994.

\bibitem{gualtieri2016high}
M.~Gualtieri, A.~ten Pas, K.~Saenko, and R.~Platt, ``{High Precision Grasp Pose
  Detection in Dense Clutter},'' in \emph{IEEE/RSJ Intl. Conf. on Intelligent
  Robots and Systems (IROS)}, 2016, pp. 598--605.

\bibitem{levine2016learning}
S.~Levine, P.~Pastor, A.~Krizhevsky, J.~Ibarz, and D.~Quillen, ``{Learning
  Hand-Eye Coordination for Robotic Grasping with Deep Learning and Large-Scale
  Data Collection},'' \emph{The International Journal of Robotics Research}, p.
  0278364917710318, 2016.

\bibitem{mahler2017dex}
J.~Mahler, J.~Liang, S.~Niyaz, M.~Laskey, R.~Doan, X.~Liu, J.~A. Ojea, and
  K.~Goldberg, ``{Dex-Net 2.0: Deep Learning to Plan Robust Grasps with
  Synthetic Point Clouds and Analytic Grasp Metrics},'' in \emph{Robotics
  Science and Systems (RSS)}, 2017.

\bibitem{johns2016deep}
E.~Johns, S.~Leutenegger, and A.~J. Davison, ``{Deep Learning a Grasp Function
  for Grasping under Gripper Pose Uncertainty},'' in \emph{IEEE/RSJ Intl. Conf.
  on Intelligent Robots and Systems (IROS)}, 2016, pp. 4461--4468.

\bibitem{varley2015generating}
J.~Varley, J.~Weisz, J.~Weiss, and P.~Allen, ``{Generating Multi-Fingered
  Robotic Grasps via Deep Learning},'' in \emph{IEEE/RSJ Intl. Conf. on
  Intelligent Robots and Systems (IROS)}, 2015, pp. 4415--4420.

\bibitem{redmon2015real}
J.~Redmon and A.~Angelova, ``{Real-Time Grasp Detection using Convolutional
  Neural Networks},'' in \emph{IEEE Intl. Conf. on Robotics and Automation
  (ICRA)}, 2015, pp. 1316--1322.

\bibitem{kumra2016robotic}
S.~Kumra and C.~Kanan, ``{Robotic Grasp Detection using Deep Convolutional
  Neural Networks},'' in \emph{IEEE/RSJ Intl. Conf. on Intelligent Robots and
  Systems (IROS)}, 2017.

\bibitem{veres2017modeling}
M.~Veres, M.~Moussa, and G.~W. Taylor, ``{Modeling Grasp Motor Imagery through
  Deep Conditional Generative Models},'' \emph{IEEE Robotics and Automation
  Letters}, vol.~2, no.~2, pp. 757--764, 2017.

\bibitem{kappler2015leveraging}
D.~Kappler, J.~Bohg, and S.~Schaal, ``{Leveraging Big Data for Grasp
  Planning},'' in \emph{IEEE Intl. Conf. on Robotics and Automation (ICRA)},
  2015, pp. 4304--4311.

\bibitem{lu2017grasp}
Q.~Lu, K.~Chenna, B.~Sundaralingam, and T.~Hermans, ``{Planning Multi-Fingered
  Grasps as Probabilistic Inference in a Learned Deep Network},'' in \emph{Int.
  Symp. on Robotics Research}, 2017.

\bibitem{zhou6dof}
Y.~Zhou and K.~Hauser, ``{6DOF Grasp Planning by Optimizing a Deep Learning
  Scoring Function},'' in \emph{Robotics: Science and Systems (RSS) Workshop on
  Revisiting Contact - Turning a Problem into a Solution}, 2017.

\bibitem{lu2019grasp}
Q.~Lu and T.~Hermans, ``{Modeling Grasp Type Improves Learning-Based Grasp
  Planning},'' 2019.

\bibitem{bishop1994mixture}
C.~M. Bishop, ``Mixture density networks,'' \emph{Technical report, Citeseer},
  1994.

\bibitem{Kendall2017}
A.~Kendall and Y.~Gal, ``{What uncertainties do we need in Bayesian deep
  learning for computer vision?}'' in \emph{Advances in Neural Information
  Processing Systems}, 2017, pp. 5575--5585.

\bibitem{byrd1995limited}
R.~H. Byrd, P.~Lu, J.~Nocedal, and C.~Zhu, ``{A limited memory algorithm for
  bound constrained optimization},'' \emph{SIAM Journal on Scientific
  Computing}, vol.~16, no.~5, pp. 1190--1208, 1995.

\bibitem{zhu1997algorithm}
C.~Zhu, R.~H. Byrd, P.~Lu, and J.~Nocedal, ``{Algorithm 778: L-BFGS-B: Fortran
  subroutines for large-scale bound-constrained optimization},'' \emph{ACM
  Trans. on Mathematical Software}, vol.~23, no.~4, pp. 550--560, 1997.

\bibitem{Fischler1981}
\BIBentryALTinterwordspacing
M.~A. Fischler and R.~C. Bolles, ``{Random Sample Consensus: A Paradigm for
  Model Fitting with Applicatlons to Image Analysis and Automated
  Cartography},'' \emph{Communications of the ACM}, vol.~24, no.~6, pp. 381 --
  395, 1981. [Online]. Available: \url{http://dx.doi.org/10.1145/358669.358692}
\BIBentrySTDinterwordspacing

\bibitem{hermans-icra2013}
T.~Hermans, J.~M. Rehg, and A.~F. Bobick, ``{Decoupling Behavior, Perception,
  and Control for Autonomous Learning of Affordances},'' in \emph{{IEEE
  International Conference on Robotics and Automation}}, 2013.

\bibitem{miller2003automatic}
A.~T. Miller, S.~Knoop, H.~I. Christensen, and P.~K. Allen, ``{Automatic grasp
  planning using shape primitives},'' in \emph{IEEE Intl. Conf. on Robotics and
  Automation (ICRA)}, vol.~2, 2003, pp. 1824--1829.

\bibitem{singh2014bigbird}
A.~Singh, J.~Sha, K.~S. Narayan, T.~Achim, and P.~Abbeel, ``{Bigbird: A
  Large-Scale 3D Database of Object Instances},'' in \emph{IEEE Intl. Conf. on
  Robotics and Automation (ICRA)}, 2014, pp. 509--516.

\bibitem{brock2016generative}
A.~Brock, T.~Lim, J.~M. Ritchie, and N.~Weston, ``Generative and discriminative
  voxel modeling with convolutional neural networks,'' \emph{arXiv preprint
  arXiv:1608.04236}, 2016.

\bibitem{Kappler2015}
\BIBentryALTinterwordspacing
D.~Kappler, J.~Bohg, and S.~Schaal, ``{Leveraging big data for grasp
  planning},'' in \emph{IEEE Intl. Conf. on Robotics and Automation (ICRA)},
  2015, pp. 4304--4311. [Online]. Available:
  \url{http://ieeexplore.ieee.org/document/7139793/}
\BIBentrySTDinterwordspacing

\bibitem{calli2015benchmarking}
B.~Calli, A.~Singh, A.~Walsman, S.~Srinivasa, P.~Abbeel, and A.~M. Dollar,
  ``{The YCB Object and Model Set: Towards Common Benchmarks for Manipulation
  Research},'' in \emph{International Conference on Advanced Robotics (ICAR)},
  2015, pp. 510--517.

\end{thebibliography}

%

\end{document}